\documentclass[wcp]{jmlr}

\usepackage{longtable}% for long tables

\usepackage{booktabs}
\usepackage{lineno}
\makeatletter
\let\Ginclude@graphics\@org@Ginclude@graphics 
\makeatother

\jmlrvolume{222}
\jmlryear{2023}
\jmlrworkshop{ACML 2023}

\title[Degree-based stratification of nodes in\\ Graph Neural Networks]{Degree-based stratification of nodes in\\  Graph Neural Networks}
 \author{\Name{Ameen Ali} \Email{ameenali@mail.tau.ac.il} \\
 \Name{Lior Wolf} \Email{wolf@mail.tau.ac.il}\\
 \addr School of Computer Science, Tel Aviv, Israel
 \AND
 \Name{Hakan Cevikalp} \Email{hakan.cevikalp@gmail.com}\\
\addr MLCV Laboratory, Eskisehir Osmangazi University, Eskisehir, Turkey
}

\editors{Berrin Yan{\i}ko\u{g}lu and Wray Buntine}

\begin{document}

\maketitle
\begin{abstract}
Despite much research, Graph Neural Networks (GNNs) still do not display the favorable scaling properties of other deep neural networks such as Convolutional Neural Networks and Transformers. Previous work has identified issues such as oversmoothing of the latent representation and have suggested solutions such as skip connections and sophisticated normalization schemes. Here, we propose a different approach that is based on a stratification of the graph nodes. We provide motivation that the nodes in a graph can be stratified into those with a low degree and those with a high degree and that the two groups are likely to behave differently. Based on this motivation, we modify the Graph Neural Network (GNN) architecture so that the weight matrices are learned, separately, for the nodes in each group. This simple-to-implement modification seems to improve performance across datasets and GNN methods. To verify that this increase in performance is not only due to the added capacity, we also perform the same modification for random splits of the nodes, which does not lead to any improvement. 
\end{abstract}
\section{Introduction}

Graph Neural Networks (GNNs) provide an effective way to learn over graphs, which is extremely applicable in domains as diverse as hardware design, economy, medicine, recommendation systems, and physics~\citep{zhou2020graph}. However, unlike other neural models, this success does not increase with depth. In fact, in most cases, shallow GNNs outperform deeper ones, cf.~\citep{guo2023contranorm}.

To overcome the paradoxical behavior of depth, some contributions try to overcome oversmoothing, the phenomenon in which node embeddings become too similar with depth \citep{wu2022non,rusch2023survey}. 
One approach to this is to incorporate skip connections \citep{xu2021optimization,xu2020neural}. Such connections are used, for example, in Graph Attention Networks (GATs)~\citep{velivckovic2017graph} and Graph Convolutional Networks with attention mechanisms (GCANs)~\citep{guo-etal-2019-attention}. Another approach is to apply specific forms for normalization, e.g., PairNorm \citep{zhao2019pairnorm}, in which pairwise node embeddings in every layer are prevented from becoming too similar, and the recent ContraNorm \citep{guo2023contranorm}, which achieves the same goal by applying a shattering operator to the latent representation. 

In this work, we study a different way to add capacity to GNNs. Instead of trying to add depth, we stratify the nodes by their graph degrees and learn a separate weight matrix for each group. Specifically, we split the nodes of the graph into low- and high-degree nodes and learn two sets of weights (one weight matrix per layer, per group). 

Recall that a node's degree is the number of nodes that are connected to it and that this is entirely dependent on the input adjacency matrix. We motivate our method by presenting evidence that the low-degree nodes are harder to classify and that they tend to lie on the decision boundaries. We argue that high-degree nodes, which tend to lie in uniform areas, are less affected by the representation collapse caused by oversmoothing. On the other hand, low-degree nodes require more careful consideration, which is obtained by assigning a specified weight matrix to handle these nodes.

By assigning one weight matrix to low-degree nodes, and another to high-degree nodes, we double the capacity of the GNN. This, as we demonstrate, is helpful across datasets and for multiple GNN architectures. However, as our ablation shows, adding this capacity by randomly splitting the nodes into two groups is detrimental to the overall accuracy, and performs worse than the original GNN. Our work, therefore, demonstrates the need to apply GNNs message passing in a way that takes into account the degree of each node.

\section{Related work}

Graph Neural Networks (GNNs) have recently attracted increasing interest as a powerful framework for learning graph-structured data. Inspired by the success of Convolutional Neural Networks (CNNs), GNNs aim to extend the idea of convolutions to handle non-grid data (graphs) by leveraging the graph structure. GNNs have been shown to learn expressive node/graph representations \citep{kipf2016semi,hamilton2017inductive,xu2018powerful} and have achieved state-of-the-art results over various graph tasks \citep{velivckovic2017graph,bojchevski2017deep,chen2020simple,gasteiger2019diffusion}.

Weight sharing is a fundamental concept that is employed in various neural network architectures, including CNNs \citep{lecun1998gradient} and GNNs \citep{hamilton2017inductive}. In these models, weight sharing refers to the practice of using the same set of weights across different regions or nodes to capture shared patterns and achieve parameter efficiency. By sharing weights, the model can generalize across regions or nodes, effectively capturing common features and relationships. In addition to weight sharing, recent research has explored other techniques to enhance the expressive power of GNNs. Attention mechanisms have gained significant interest in the field of graph representation learning. Graph Attention Networks (GAT) \citep{velivckovic2017graph} proposed using attention mechanisms for neighbors aggregation, allowing nodes to attend to their neighbors with varying weights. While weight sharing and attention mechanisms have contributed to the success of GCNs, there is ongoing research on further enhancing their capabilities. \citep{zhang2022graph} proposed Graph-adaptive Rectified Linear Unit (GReLU), a topology-aware parametric activation function that incorporates neighborhood information in a novel and efficient way, improving GNN performance across various tasks.  \citep{wang2022powerful}  introduces a novel propagation mechanism in Graph Convolutional Networks (GCNs) that adaptively adjusts the propagation and aggregation process based on the homophily or heterophily between node pairs. \citep{kim2022find} presents an improved model for noisy graphs, utilizing two attention forms compatible with a self-supervised task to predict edges and encode relationships, resulting in more expressive attention in distinguishing mislinked neighbors. In \citep{chen2023exploiting} the authors argue that the traditional perspective on inter-class edges and existing metrics cannot fully explain the performance of Graph Neural Networks (GNNs) on heterophilic graphs, and propose a new metric based on von Neumann entropy to re-examine the heterophily problem. They also introduce a Conv-Agnostic GNN framework (CAGNNs) that enhances GNN performance on heterophilic datasets by learning the neighbor effect for each node, achieving significant improvements in performance compared to GIN, GAT, and GCN models on various benchmark datasets. \citep{wu2019net} addresses the limitations of existing graph neural networks by proposing DEMO-Net, a degree-specific graph neural network that incorporates seed-oriented, degree-aware, and order-free graph convolution properties, along with multi-task learning methods and a novel graph-level pooling scheme, achieving superior performance on node and graph classification benchmarks. Inspired by the Transformer architecture, Graph Transformers \citep{yun2019graph,hu2020heterogeneous,ying2021transformers,dwivedi2020generalization,dwivedi2021graph}  incorporate a self-attention mechanism with multiple heads to assign varying weights to neighbor nodes, enabling effective information aggregation and representation learning. 

\section{Motivation}
\label{sec:analysis}

We build our proposed method on the cluster assumption of \citep{Chapel}  which states that the decision boundaries should not cross high-density regions and should lie in low-density regions. To this end, we split the node degrees into two separate sets according to high- and low-degree nodes, and create two graphs, by using each. We treat these two graphs as if they were two different graphs and obtain weight matrices for both, which, as we demonstrate, improves classification accuracy.

The data samples corresponding to the nodes with high degrees lie in high-density cluster regions surrounded by the neighboring data samples from the same class categories. On the other hand, the samples from the nodes with lower degrees typically lie in the critical regions where the classes overlap with each other, i.e., inter-class decision boundaries\footnote{Most of the node embeddings with lower node degrees lie closer to decision boundaries and the remaining ones lie in isolated regions far from both the decision boundaries and high-density cluster regions.}. These nodes are critical for successful classification and the propagation of information among them must be slower so that the nodes from different class categories do not collapse to similar points and hence can be separated from each other. To verify our claims, we classified the nodes with lower and higher degrees separately, using the different GNN baseline methods in Table \ref{table:results_seperate}. As can be seen from the table, it is evident that the classification accuracies achieved by the baseline models exhibit a notable disparity between nodes with lower degrees and those with higher degrees across all three datasets. Specifically, the accuracies attained for nodes with lower degrees are considerably inferior in comparison to the accuracies observed for nodes with higher degrees. %(We can also put tsne plots if we obtained meaningful plots)

\begin{figure}
    \centering
    \begin{tabular}{c@{}cc}
    \includegraphics[width=0.45\textwidth]{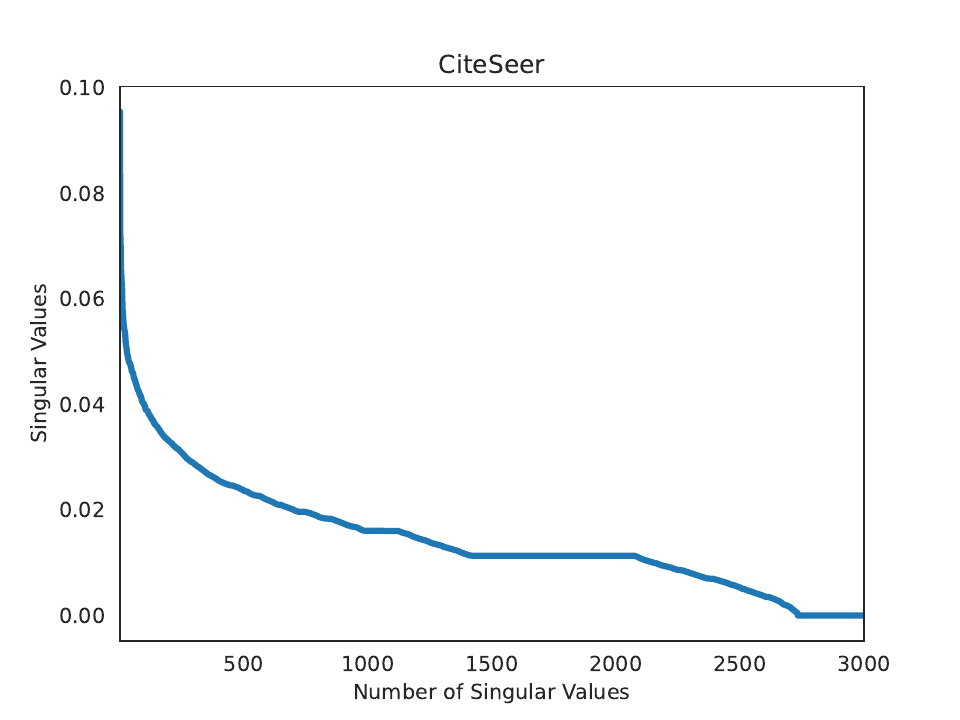} &\includegraphics[width=0.45\textwidth]{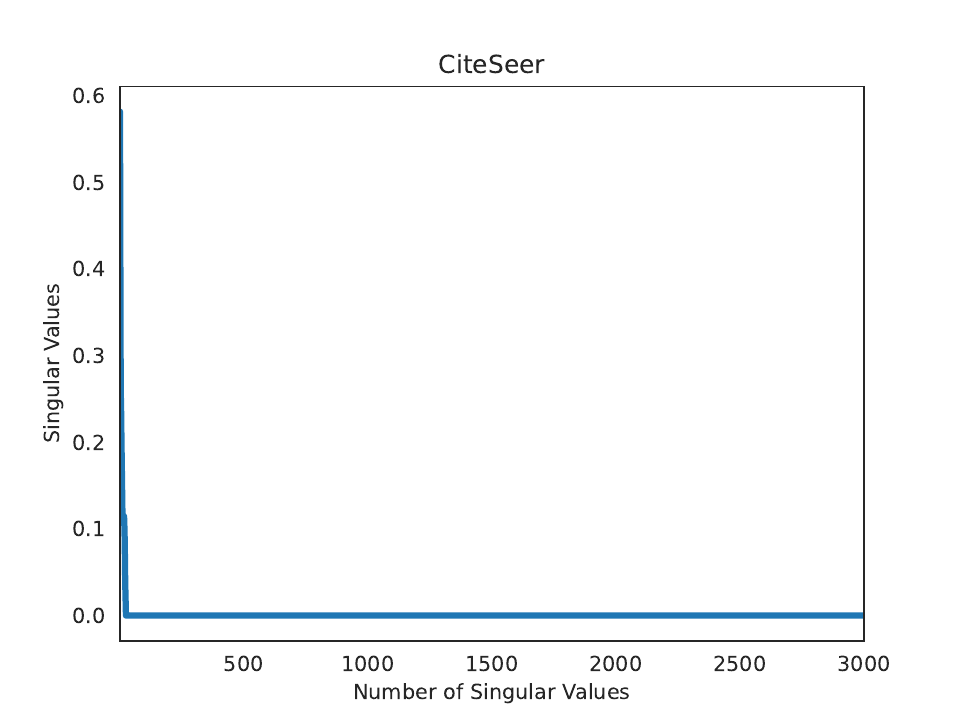} \\
    \includegraphics[width=0.45\textwidth]{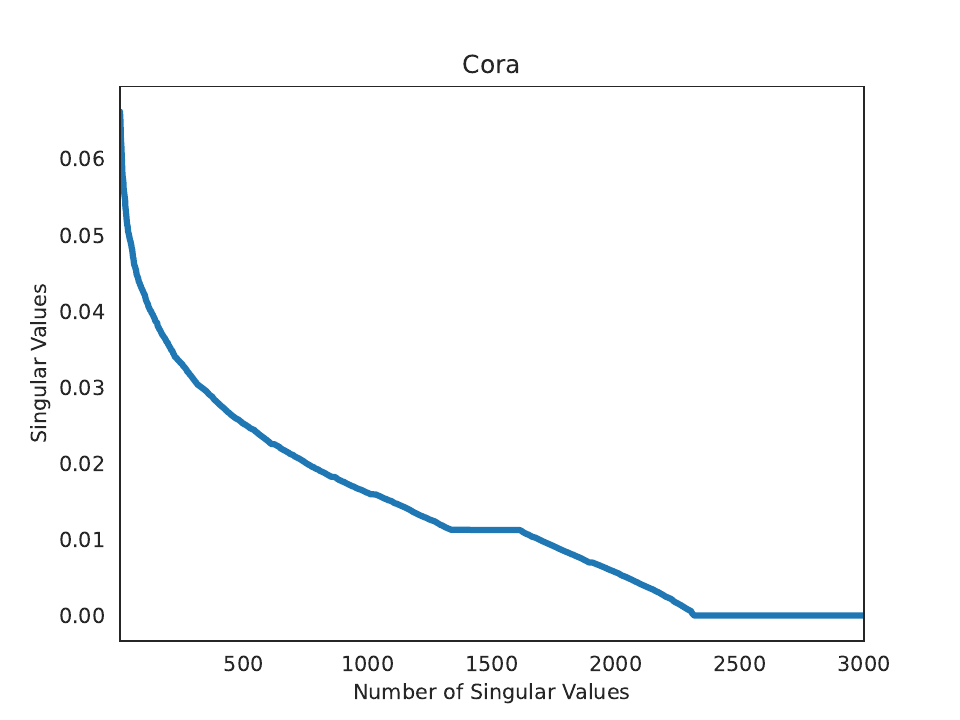} &\includegraphics[width=0.45\textwidth]{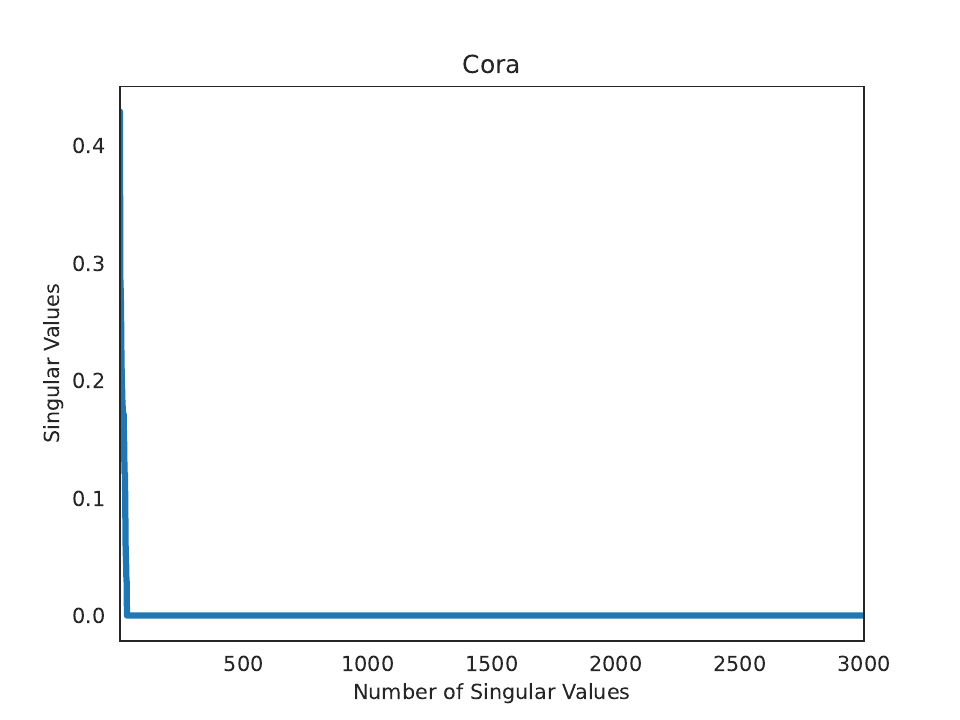} \\
    \includegraphics[width=0.45\textwidth]{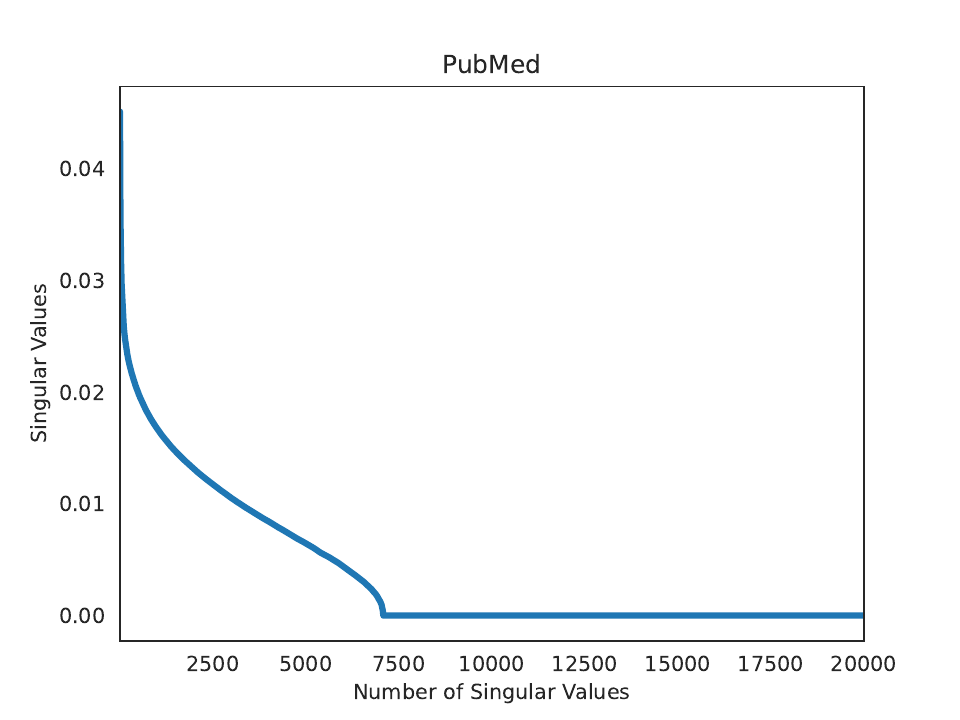} &\includegraphics[width=0.45\textwidth]{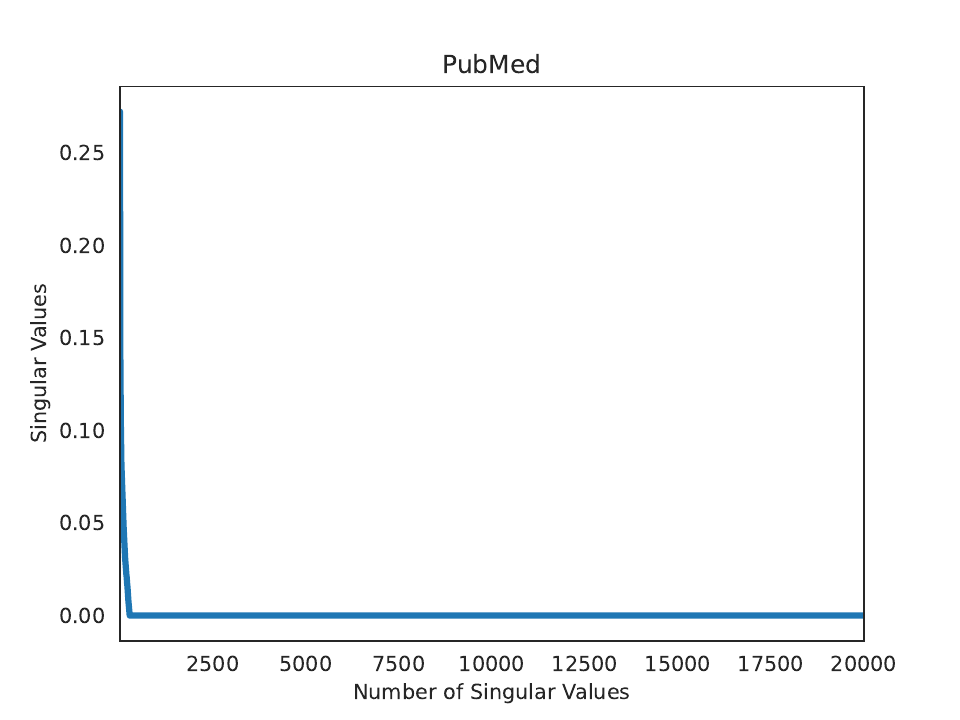} \\
    (a) & (b)\\
\end{tabular}
\caption{Spectrum of the renormalized adjacency over the low-degree nodes graph (a) and high-degree nodes graph (b).}
\label{fig:spectrum}
\end{figure}

In graph convolutional networks, high-degree nodes have a larger number of connections or neighbors, therefore information spreads more quickly and widely through them. On the other hand, low-degree nodes have fewer connections and may not be directly connected to as many nodes as high-degree nodes. Therefore, information originating from low-degree nodes may take longer to propagate through the network, since it needs to traverse a longer path through intermediate nodes to reach other parts of the graph. This is also reflected in the spectrum plots of the adjacency matrices obtained from the graphs using lower and higher node degrees given in Fig. \ref{fig:spectrum}. As indicated in \citep{Yang}, if eigenvalues have similar magnitudes, the resulting graph nodes are more robust to over-smoothing, whereas the node embeddings become similar very rapidly for the graphs with an adjacency matrix containing dissimilar eigenvalues where the majority of the eigenvalues are insignificant (nearly zero).  As a result, when we use the full graph structure without splitting the node degrees, the node embeddings coming from the opposing classes near the decision boundaries become rapidly similar. Therefore, the decision boundaries merge and the node embeddings become nonseparable. To avoid this, we split the nodes into lower- and higher-degree nodes, treat them as two different graphs and learn two weight matrices, for each. This way, information will propagate more slowly among nodes with lower degrees, since their eigenvalue spectrum is more uniform. As a result, the model will be more robust to over-smoothing. 

\section{Method}
In this section, we describe the proposed method of using different weight matrices for different node degree groups in a Graph Neural Network. The approach aims to enhance the learning capability of the GNN by considering the varying degrees. By assigning separate weight matrices to different node degrees, the model can capture degree-specific patterns, which improve its expressivity.

\subsection{Preliminaries} We begin by providing a brief overview of the general GNN architecture that serves as the foundation for our proposed approach.
A GNN operates on a graph represented by an adjacency matrix $A \in \mathbb{R}^{n\times n}$ where $n$ is the total number of nodes in the given graph, and a node feature matrix $X \in \mathbb{R}^{n \times d}$ where $d$ is the feature dimensionality of each node vector. At each layer, $i$, the GNN updates the node representations based on their neighborhood information and propagates the information through the graph. The specific update rules vary depending on GNN architecture. The updating rule for a GCN model \citep{kipf2016semi} can be formulated as follows:

Let $H^i \in \mathbb{R}^{n \times F(i)}$ denote the hidden representations of the different nodes of a certain graph at the $i$-th layer of the GCN, where $F(i)$ denotes the hidden dimension at layer $i$, for $i=0$ we have $F(i) = F(0) = d$, the propagation rule of the GCN can be defined as follows:
\begin{center}
    \begin{equation}
        H^{i+1} = f(A , H^i , W^i) = \sigma(D^{-\frac{1}{2}} \cdot A \cdot D^{-\frac{1}{2}} \cdot H^i \cdot W^i )\,,
    \end{equation}
\end{center}
where $W^i \in \mathbb{R}^{F(i)\times F(i+1)}$ is the weight matrix at the $i$-th layer, $\sigma(\cdot)$ is the activation function, like ReLU/ELU and $D \in \mathbb{R}^{n\times n}$ is the degree matrix obtained by $D_{v,u} = \sum_{u}A_{v,u}$.

Similarly, the propagation rule of Graph Attention Networks (GATs) \citep{velivckovic2017graph} takes a similar form, except that it involves first computing attention coefficients between each node and its neighbors. For node $k$ and its neighbor $j$, the attention coefficient $e_{kj}^{i}$ at layer $i$ is computed as follows:

\begin{center}
    \begin{equation}
    e_{kj}^{i} = LeakyReLU({a^i}^\top [W^i h_k^i || W^i h_j^i])\,,
    \end{equation}
\end{center}
where $h_k^i , h_j^i \in \mathbb{R}^{F(i) , }$ are the  hidden representations of nodes $k$ and $j$ at layer $i$. $W^i \in \mathbb{R}^{m \times F(i)}$ is the weight matrix at layer $i$ which is used to transform the node features, $a^i \in \mathbb{R}^{2\times m, }$ is the attention parameter vector used in the shared attention mechanism at layer $i$ and $||$ denotes the concatenation operator. $e^i_{kj} \in \mathbb{R}$ representing the attention coefficient; the attention scores are subsequently obtained by : 
\begin{center}
    \begin{equation}
        \alpha_{kj}^i = \frac{exp(e^i_{kj})}{\sum_{m \in \mathcal{N}(k)} exp(e^i_{km})}
    \end{equation}
\end{center}

The aggregating step is based on these scores rather than uniform aggregating as in GCNs, which is formulated in the following equation :
\begin{center}
    \begin{equation}
        h^{i}_k = \sigma(\sum_{j \in \mathcal{N}(k)} \alpha_{kj}^i \cdot h_j^{i-1})
    \end{equation}
\end{center}

Lastly, a Graph-SAGE model operates on a graph by aggregating and updating information from up to $k$-hop neighborhoods for each node. The model follows a cyclic process for each node in the graph, where it gathers information from the node's $k$-hop neighborhood, aggregates that information and then updates the node's representation. This process is repeated iteratively to refine the node representations. The aggregation \& updating cycle is formulated as follows: for each node $v$, aggregate the representations from its $k$-hop neighborhood $\mathcal{N}_k(v)$:
\begin{center}
    \begin{equation}
        h^k_{\mathcal{N}(v)} = AGGREGATE(h^{k-1}_{u} | d(u , v) = k)
    \end{equation}
    \begin{equation}
        h^k_v = \sigma(W^k \cdot CONCAT(h^{k-1}_v , h^k_{\mathcal{N}(v)}))
    \end{equation}
\end{center}
Where $h^0_v$ is the initial node feature representation, $K$ is the depth hyperparameter, $k \leq K$ is the iteration step, $W^k$ is a weight matrix for iteration $k$, $d(u,v)$ represents the shortest path distance between nodes $u$ and $v$ and $AGGREGATE$ is a placeholder for any differentiable function.

The above formulations use a single-weight matrix $W^i$ for each layer $i$ to capture the interactions between the different nodes. In our proposed approach, we aim to extend the standard formulation by incorporating different weight matrices for each group of node degrees.

\subsection{Degree-Specific Weight Matrices}

In our approach, we propose extending the standard GNNs with the use of different weight matrices for each group of node degrees. The motivation behind this is to account for the variation of the properties between the groups, since nodes with different degrees may contribute differently to the overall task at hand, see Sec.~\ref{sec:analysis}.

For each layer in the GNN, we introduce a set of weight matrices, each corresponding to a specific group of node degrees. For maximal simplicity, we set the number of weight matrices to two, accounting for nodes with high and low degrees. In preliminary experiments, even setting one weight matrix for each degree that is present in the graph works well. However, this results in a much larger model that makes an apple-to-apple comparison hard. Yet, only two degree-specific weight matrices can already capture the degree-specialized relationships and enable the model to learn different patterns associated with nodes of different degrees.

In order to incorporate the degree-specific weight matrices, we modify the updating rule of the GNN to consider node degrees during the propagation step. The specific formulation depends on the GNN architecture being used. For instance, for a GCN model, the update rule at the $i$-th layer can be modified as follows :

\begin{center}
    \begin{equation}
        H^{i+1}_{low} = f(A , H^i , W^i_{low}) = \sigma(D^{-\frac{1}{2}} \cdot A \cdot D^{-\frac{1}{2}} \cdot H^i \cdot W^i_{low} )\,,
    \end{equation}
    \begin{equation}
        H^{i+1}_{high} = f(A , H^i , W^i_{high}) = \sigma(D^{-\frac{1}{2}} \cdot A \cdot D^{-\frac{1}{2}} \cdot H^i \cdot W^i_{high} )\,,
    \end{equation}
    \begin{equation}
        H^{i+1}[k,:] = \begin{cases}
            H^{i+1}_{low}[k , :], & \text{if $Deg[k] \leq \theta$} \\
            H^{i+1}_{high}[k, : ], & \text{if $Deg[k] > \theta$}\,,
        \end{cases}
    \end{equation}
\end{center}
where $H_{low}^{i+1} \in \mathbb{R}^{n \times F(i+1)}, H_{high}^{i+1}\in \mathbb{R}^{n \times F(i+1)}$ represents the hidden representation at layer $i$ as computed for low and high degrees nodes, respectively, (it is written this way to simplify notation; in practice, we discard half of these computations) $Deg \in \mathbb{R}^{n \times 1}$ is a vector that represents the degree of the corresponding node, square brackets are used to indicate a selection of the column or element that corresponds to a specific node, and $\theta$ is the threshold used to split nodes into low/high degree sets.

Similarly, in a GAT, we modify the attention coefficient calculation in the following way

\begin{center}
    \begin{equation}
        e_{kj}^{i} = \begin{cases}
         LeakyReLU({a^i}_{low}^\top [W^i_{low} h_k^i || W^i_{low} h_j^i]), & \text{if $Deg[k] \leq \theta$} \\
         LeakyReLU({a^i}_{high}^\top [W^i_{high} h_k^i || W^i_{high} h_j^i]), & \text{if $Deg[k] > \theta$} \\
        \end{cases}
    \end{equation}
\end{center}
Where $W^i_{low} , W^i_{high} \in \mathbb{R}^{m \times F(i)}$ , ${a^i}_{low}^\top , {a^i}_{high}^\top \in \mathbb{R}^{2 \times m}$ are the weight matrices and attention parameters vectors at layer $i$, respectively, for both low and high degrees.

Lastly, in graphSAGE we modify the update step in each iteration of the cyclic process of aggregation and updating as follows:
\begin{center}
    \begin{equation}
        h^k_v = \begin{cases}
                    \sigma(W^k_{low} \cdot CONCAT(h^{k-1}_v , h^k_{\mathcal{N}(v)})), & \text{if $Deg[v] \leq \theta$} \\
                    \sigma(W^k_{high} \cdot CONCAT(h^{k-1}_v , h^k_{\mathcal{N}(v)})), & \text{if $Deg[v] > \theta$} \\
        \end{cases}
    \end{equation}
\end{center}

\section{Experiments}
In this section, we present the experimental setup, datasets, and the results obtained from evaluating the proposed approach of using different weight matrices for each group of node degrees in different types of Graph Neural Networks (GNNs).
\subsection{Datasets}
We have evaluated our approach over three commonly used node classification datasets: Cora, Citeseer, and PubMed. These datasets represent citation networks where the nodes correspond to documents and edges represent citation relationships. Each node has associated textual features, and the task is to classify the nodes into predefined categories based on their features and connectivity.
The Cora and CiteSeer datasets consist of graphs representing academic papers, where nodes represent papers and the edges represent citation relationships between them, the nodes are assigned to a specific set of academic topics, and the task is to classify each node (paper) into different classes based on their topics. The Cora dataset contains $2708$ nodes with a total of 5278 edges, where each node is represented by a $1433$ dimensional feature vector; the total number of classes is $7$. The CiteSeer dataset contains $3327$ total nodes and $4552$ edges, with each node represented by a $3703$ dimensional feature vector, and the total number of classes in this dataset is 6. The PubMed dataset contains a total of $19717$ nodes (scientific publications), which are categorized into $3$ classes, with a total of $44338$ edges (connections between the publications); each node in the graph is represented by a TF/IDF word vector constructed using a vocabulary of $500$ unique words.
\subsection{Models} In our experiments, we verify the effectiveness of our proposed approach of learning a different weight matrix per each set of node degrees on several graph-learning-based models commonly used in graph analysis. Specifically, we implemented our method on three graph representation learning models: Graph Convolutional Neural Network (GCN), Graph Attention Network (GAT), and GraphSAGE.
By including these three baseline methods, we aim to evaluate the performance of our proposed approach comprehensively and assess its effectiveness. 
\subsection{Implementation Details}
We implemented our proposed approach using the PyTorch library~\citep{NEURIPS2019_9015}. The models were trained using an Adam optimizer with a learning rate of 0.001 and a weight decay of 5e-4. In all of the experiments, we used a two-layer architecture, with a hidden dimension of 32 for the first and second layers. The ReLU activation function was applied after each layer, and a softmax function was used for the final classification layer.
The parameter $\theta$ is a hyperparameter used to split the nodes into two groups, those of low degree and those of high degree. To determine $\theta$, we first calculate the degree of each node in the graph and create a histogram of node degrees. We then automatically identify a threshold that divides the nodes into low-degree and high-degree groups by analyzing the histogram for a significant gap in the degree distribution. This is done using the conventional Otsu’s method \cite{otsu1979threshold}. In our experiments, we use $\theta=2,3,5$ for Cora, CiteSeer, and PubMed datasets respectively.
\begin{table}
\caption{Classification accuracy of the baseline GCN and our method over high- and low-degree nodes for different datasets.}
\smallskip
\smallskip
\centering
    \begin{tabular}{llcc}
    \toprule
    % \multicolumn{9}{c} {Node Classification} \\
    % \midrule
    \textbf{Dataset} & \textbf{Variant} & \textbf{Low-Degree}  & \textbf{High-Degree} 
    \\
     &  & \textbf{Nodes}  & \textbf{Nodes} 
    \\
    \midrule
    Cora &Baseline & 0.775 & 0.809   \\
    Cora &Ours & \textbf{0.798}& \textbf{0.831} \\

    \midrule
    CiteSeer &Baseline & 0.612& 0.904   \\
    CiteSeer &Ours & \textbf{0.661} & \textbf{0.965} \\

    \midrule
    PubMed &Baseline & 0.737& 0.775    \\
    PubMed &Ours& \textbf{0.758}& \textbf{0.801}  \\
    \bottomrule
    \label{table:results_seperate}
    \end{tabular}
\end{table}

\begin{table}
\caption{Classification accuracy over the test set of different datasets, results are averaged across 10 different runs, we used the default hyperparameters of the \citep{guo2023contranorm} method.}
\smallskip
\smallskip
    \begin{tabular}{@{}l@{~~}l@{~}ccc@{}}
    \toprule
    % \multicolumn{9}{c} {Node Classification} \\
    % \midrule
    \textbf{Dataset} & \textbf{Variant} & \textbf{GCN}  & \textbf{GAT} & \textbf{Graph SAGE}
    \\
    \midrule
    Cora &Baseline & 0.7909 $\pm$ 0.012& 0.7885 $\pm$ 00072 & 0.7256 $\pm$ 0.006  \\
    Cora &Ours & \textbf{0.8035 $\pm$ 0.003}& \textbf{0.8109 $\pm$ 0.004} & \textbf{0.7463 $\pm$ 0.10}\\
    Cora & Ablation (rand division)& 0.7749 $\pm$ 0.015& 0.7710 $\pm$ 0.013& 0.7067 $\pm$ 0.007    \\

    \midrule
    CiteSeer &Baseline & 0.6308 $\pm$ 0.010 & 0.6578 $\pm$ 0.0054 & 0.6206 $\pm$ 0.004   \\
    CiteSeer &Ours & \textbf{0.6491 $\pm$ 0.009}& \textbf{0.6787 $\pm$ 0.007} & \textbf{0.6431 $\pm$ 0.010}\\
        CiteSeer & Ablation (rand division)& 0.6231 $\pm$ 0.012& 0.6452 $\pm$ 0.017& 0.6182 $\pm$ 0.009   \\

    \midrule
    PubMed &Baseline & 0.7633 $\pm$ 0.002& 0.7626 $\pm$ 0.005& 0.7402 $\pm$ 0.009    \\
    PubMed &Ours& \textbf{0.7739 $\pm$ 0.005}& \textbf{0.7724 $\pm$ 0.003}  & \textbf{0.7512 $\pm$ 0.012} \\
        PubMed & Ablation (rand division)& 0.7523 $\pm$ 0.012& 0.7611 $\pm$ 0.019 &0.7388 $\pm$ 0.023    \\
    \bottomrule
    \label{table:results}
    \end{tabular}
\end{table}

\subsection{Results}
We conducted an initial evaluation to assess the comparative performance of our method against the baseline GCN models for classifying low-degree and high-degree nodes separately. Table \ref{table:results_seperate} presents the results. As can be seen in the results, low-degree nodes are harder to separate, indicating that they lie closer to the decision boundaries. It is also worth mentioning that our method consistently improves all other models across all tested cases.

In Table~\ref{table:results}, we present the results of our proposed approach along with the results of the vanilla GNN models from each architecture. From the results, we observe that incorporating different weight matrices based on a node degree threshold leads to improved performance compared to the baseline model (which has one shared weight matrix per layer) across all GNN variants and datasets. This suggests that considering the node degree threshold for weight assignment can capture meaningful variations in the graph structure.

\subsection{Ablation study}
\begin{figure*}[!t]
\hfill
\subfigure[]{\includegraphics[width=0.49\textwidth]{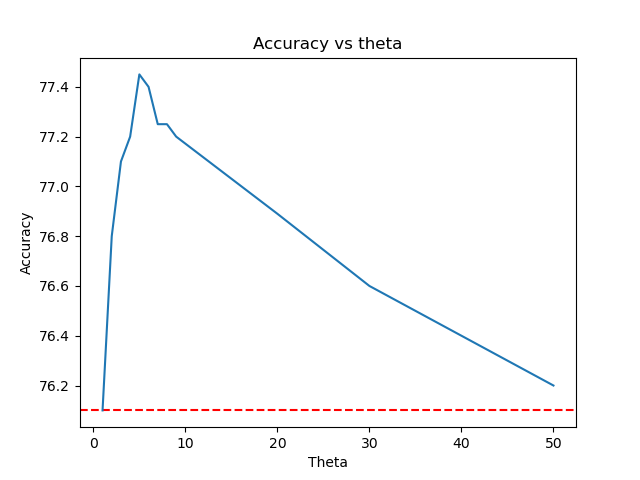}}
\hfill
\subfigure[]{\includegraphics[width=0.49\textwidth]{figures/1.png}}
\hfill
\caption{(a) the effect of the parameter $\theta$ on the accuracy for our variant of GCN applied to PubMed dataset. The red line is the baseline accuracy and the blue is our result. (b) the distribution of node degrees in the PubMed dataset.}
\label{fig:theta}
\end{figure*}

To show that the added capacity arises from the stratification by node degree and not just from having a second weight matrix per layer, we conduct an ablation study. In this study, we divide the nodes randomly into two groups, whose sizes are identical to what would have been obtained had the division been based on the degree.  Furthermore, we conducted an additional random division of nodes into two groups. As can be seen in Table~\ref{table:results}, the added capacity, if assigned randomly, (slightly) hurts performance instead of improving it. Additionally, in order to study the effect of different $\theta$ values on the performance, we have applied an ablation study over the PubMed dataset, the results can be seen in Fig.~\ref{fig:theta} .

 \section{Conclusions}
The message-passing mechanism employed by GNNs aggregates the information from multiple neighbors in a way that is agnostic to the local topography of the graph, except for the set of neighbors. In this work, we rely on the simplest node property (the node's degree) and the simplest way to create a topology-dependent treatment of nodes (separation into two groups by thresholding the degree). 

With this simple method, we are able to show a very consistent improvement in GNN performance. This is true for both the simplest graph convolutional network and for attention-based variants. Our ablation clearly demonstrates that the improvement in accuracy is not due to the increase in network capacity.

In future work, we would like to rely more heavily on the graph's topology, by encoding the local neighborhood of each node and utilizing this information to condition the message-passing process. 

\acks{This project was partly funded by a grant from the Tel Aviv University Center for AI and Data Science (TAD).}

%\acks{Acknowledgements should go at the end, before appendices and references. You can uncomment this for the camera-ready version on paper acceptance.}

%\bibliographystyle{plain}
\bibliography{acml23}

% \appendix

% \section{First Appendix}\label{apd:first}

% This is the first appendix.

% \section{Second Appendix}\label{apd:second}

% This is the second appendix.

\end{document}